\documentclass{article}
\usepackage{arxiv}
\usepackage[utf8]{inputenc} 
\usepackage[T1]{fontenc}
\usepackage{hyperref}  
\usepackage{url}  
\usepackage{booktabs}       
\usepackage{amsfonts}       
\usepackage{nicefrac}      
\usepackage{microtype}     
\usepackage{graphicx}
\usepackage{paralist}
\usepackage{amsmath, amsfonts}
\usepackage{xcolor}
\definecolor{forestgreen}{HTML}{32A852}
\usepackage{amssymb}
\usepackage{amsthm}
\usepackage{mathtools}
\usepackage{comment}
\usepackage{caption}
\usepackage{subcaption}
\usepackage{bm}
\usepackage{tabularx}
\usepackage{booktabs}
\usepackage{floatrow}
\usepackage{dcolumn}
\usepackage{adjustbox}
\usepackage{pgfplots}
\usepackage{appendix}
\usepackage{tabularray}
\usetikzlibrary{external}
\usepackage{placeins}
\usepackage{titlesec}
\titlespacing*{\section}{0pt}{0.5em}{0.5em}

\title{GNNsuite: A Graph Neural Network bench-marking framework for Biomedical Informatics}

\author{
 Sebesty\'en Kamp \\
  School of Informatics\\
  University of Edinburgh\\
  \texttt{sebestyen.kamp@ed.ac.uk} \\
  \And
  Giovanni Stracquadanio \\
  School of Biological Sciences\\
  University of Edinburgh\\
  \texttt{giovanni.stracquadanio@ed.ac.uk} \\
  \And
  T. Ian Simpson \\
  School of Informatics\\
  University of Edinburgh\\
  \texttt{ian.simpson@ed.ac.uk} \\
}

\DeclareMathAlphabet\mathbfcal{OMS}{cmsy}{b}{n}

\theoremstyle{plain}

\theoremstyle{definition}

\newtheorem*{example*}{Example}

\theoremstyle{remark}

\newtheorem*{remark*}{Remark}

\begin{document}

\maketitle

\begin{abstract}

We present GNN-Suite, a robust and modular framework designed to construct and benchmark Graph Neural Network (GNN) architectures for computational biology applications. GNN-Suite standardises experimentation and reproducibility using the Nextflow workflow framework for evaluating GNN model performance. To illustrate its utility, we apply the system to the identification of cancer-driver genes by constructing molecular networks from protein-protein interaction (PPI) data sourced from STRING and BioGRID and annotating these nodes with features extracted from the Pan-Cancer Analysis of Whole Genomes (PCAWG), Pathway Indicated Drivers (PID), and COSMIC Cancer Gene Census (COSMIC-CGC) repositories.\\

Our experimental design ensures a fair comparison among diverse GNN architectures, including GAT, GAT3H, GCN, GCN2, GIN, GTN, HGCN, PHGCN, and GraphSAGE and compares them to a baseline Logistic Regression (LR) model. All GNNs were adapted to a standardised two-layer configuration and trained using uniform hyperparameters (dropout = 0.2, Adam optimiser with learning rate = 0.01, and adjusted binary cross-entropy loss to mitigate class imbalance) over an 80/20 train-test split for 300 epochs. Each model was evaluated over 10 independent runs with different random seeds to ensure statistically robust performance metrics with balanced accuracy (BACC) as the primary evaluation measure. Notably, GCN2 achieved the highest BACC (0.807 ± 0.035) on a STRING-based network, although all model types showed significant improvement over a baseline logistic model, underscoring the value of network-based learning approaches over feature-only ones.\\

Our findings show how the use of a common implementation and evaluation framework for different GNN architectures can aid robust identification not only of the best model architecture, but also of the most informative ways to incorporate data. We demonstrate that the selection of suitable complementary data in network construction and annotation has important downstream consequences for performance and model selection. By making GNN-Suite publicly available, we aim to foster reproducible research and promote improved benchmarking standards for GNN methodologies in computational biology. Future work will explore additional omics datasets and further refine network architectures to enhance predictive accuracy and interpretability in complex biomedical applications.\\

Availability: https://github.com/biomedicalinformaticsgroup/gnn-suite
    
\end{abstract}

\clearpage
\section*{Introduction}
\label{sec:introduction}

Geometric Deep Learning (GDL) is a subset of Deep Learning (DL) that generalises more traditional methods to non-Euclidean, structured data \cite{bronstein_geometric_2017}. A primary application of GDL is graph neural networks (GNNs) \cite{scarselli_graph_2009} (see Appendix \ref{app:gnns} for a brief introduction to GNNs). The first GNN, the Graph Convolutional Network (GCN), was introduced by Kipf and Welling in 2017, generalising convolutional neural networks (CNNs) to graph data and achieving state-of-the-art performance on knowledge graphs and citation networks \cite{kipf_semi-supervised_2017}. Since their first introduction, GNN architectures have been extended in various ways. Graph attention networks (GATs) incorporate the attention mechanism \cite{vaswani_attention_2023} within each layer, improving node representation by dynamically weighting neighbour contributions \cite{velickovic_graph_2018}. Other notable GNNs include Graph SAmple and aggreGatE (GraphSAGE) \cite{hamilton_inductive_2018}, Graph Isomorphism Network (GIN) \cite{xu_how_2019}, and GCN2 \cite{chen_simple_2020}, all of which can be understood as instances of Message Passing Neural Networks (MPNNs)  \cite{gilmer_neural_2017}. Today, GNNs are widely used in several domains, such as social network analysis, knowledge graphs, and molecular property prediction \cite{wu_comprehensive_2021, du_machine_2024, zhou_graph_2020, Shi_2020_CVPR, reau_deeprank-gnn_2023, yasunaga_qa-gnn_2022}. In systems biology, they have been used to analyse molecular interactions and gene expression, in population genetics to identify evolutionary patterns, and in drug discovery to model drug-target interactions and discover new antibiotics \cite{schulte-sasse_integration_2021, zhang_graph_2021, stokes_deep_2020}. 

The rapid development of different GNN architectures and their burgeoning applications makes it crucial to be able to rigorously compare different techniques, since different authors implement their models with different training and evaluation procedures that make a standardised comparison difficult. Furthermore, the different architectures may exhibit dataset-specific advantages that are not immediately apparent. The need for standardised comparisons has led to the development of several benchmarking systems for GNNs.

In 2022, Dwivedi et al. introduced a standardised framework that includes real-world data from chemistry (ZINC, AQSOL), social/academic networks (OGBL-COLLAB, WikiCS), and computer vision (MNIST, CIFAR10) \cite{dwivedi_benchmarking_2022}. The platform is built upon two main libraries: PyTorch \cite{paszke_pytorch_2019} and Deep Graph Library (DGL) \cite{wang_deep_2020}. The framework has allowed researchers to test new ideas, such as data processing, aggregation functions, pooling mechanisms, and GNN designs and performance. In 2023, instead of static graphs, Zhong et al. created GNNFlow adapted for temporal GNNs, allowing for significantly faster training than prior systems \cite{zhong_gnnflow_2023}. More recently, Gong et al. proposed another benchmarking framework for static-graphs with an API interface, this time focusing on PyTorch and TensorFlow \cite{gong_gnnbench_2024}. 

In this work, we present the first GNN benchmarking framework built with the Nextflow scientific workflow system (see Appendix \ref{app:nextflow} for a brief introduction to Nextflow). Distinguishing from prior work, Nextflow is particularly popular in the biology domain for computationally driven research \cite{di_tommaso_nextflow_2017, chen_systematic_2021, van_maldegem_characterisation_2021, khozoie_scflow_2021}. It has been used for the streamlined processing of Oxford Nanopore long-read RNA-Seq data with nanoseq, the analysis of single-cell RNA sequencing via scFlow, and the automated analysis of imaging mass cytometry for tumour microenvironment studies (imcyto).

We implement a pipeline that allows for automated comparison of different GNN architectures, written with the Nextflow computational framework.  We provide the user with access to a large array of popular GNN architectures, implemented in the PyTorch Geometric (PyG) library: Graph Convolutional Networks (GCN) \cite{kipf_semi-supervised_2017}, Graph Attention Networks (GAT) \cite{velickovic_graph_2018}, Hierarchical Graph Convolutional Networks (HGCN), Parallel Hierarchical Graph Convolutional Networks (PHGCN), Graph SAmpling and aggreGatE (GraphSAGE) \cite{hamilton_inductive_2018}, Graph Transformer Networks (GTN) \cite{shi_masked_2021}, Graph Isomorphism Networks (GIN) \cite{xu_how_2019}, and Graph Convolutional Networks II (GCN2) \cite{chen_simple_2020}.

Next, to demonstrate its usefulness, we use GNN-Suite to benchmark GNN performance when classifying cancer-driver genes. This task remains critically important as the global cancer burden continues to rise. In 2020 alone, more than 19.3 million people were diagnosed with cancer, with predictions estimating a rise to close to 28.4 million a year by 2040 \cite{sung_global_2021}. Cancer is traditionally thought to develop due to the accumulation of mutations in key genes, namely cancer driver genes (CDGs) \cite{vogelstein_cancer_2013}. These drivers can be tumour suppressor genes (TSGs) or oncogenes (OGs), and when mutated, can promote cancer's growth, spread, and progression. Cancer development, however, cannot be attributed solely to individual genes, but to the interaction of these in complex networks and pathways. Aberrations in these pathways can result in cancerous phenotypes \cite{stratton_cancer_2009, hanahan_hallmarks_2011}. The lack of precise understanding of cancer genesis has hindered the development of effective therapies. Advances in high-throughput sequencing, particularly next-generation sequencing (NGS), have resulted in the accumulation of vast genomic, transcriptomic, and chemo-molecular datasets including Pan-Cancer Analysis of Whole Genomes (PCAWG) \cite{aaltonen_pan-cancer_2020}, Pathway Indicated Drivers (PID) \cite{reyna_pathway_2020}, COSMIC Cancer Gene Census (COSMIC) \cite{sondka_cosmic_2018}, GDSC \cite{yang_genomics_2013}, PubChem \cite{kim_pubchem_2021}, and ChEMBL \cite{mendez_chembl_2019}.

We use protein-protein interaction (PPI) data from the STRING \cite{szklarczyk_string_2021} and BioGRID \cite{oughtred_biogrid_2021} databases to construct protein networks, attach cancer gene association likelihoods derived from PCAWG \cite{aaltonen_pan-cancer_2020} data as node features and label known cancer drivers using gene lists from PID \cite{reyna_pathway_2020} and COSMIC \cite{sondka_cosmic_2018}. Using these data models, we use GNN-Suite to evaluate the effectiveness of various state-of-the-art GNN architectures in predicting the likelihood of a gene functioning as a cancer driver.

\section*{Methods}
\label{sec:methods}

\subsection*{Nextflow Pipeline}

The code is implemented in Nextflow (v22.10.1) to make GNN benchmarking highly modular, following the FAIR principles \cite{wilkinson_fair_2016}. This design allows for the integration and evaluation of additional GNN architectures and ensuring adaptability for future research.

The \texttt{main.nf} script defines the Nextflow workflow. This file specifies channels and processes for training GNNs, plotting metrics, and computing evaluation statistics. A \texttt{base.config} file includes basic workflow information, default execution settings and compute-resource variables. Additionally, it contains multiple profile configurations for various environments such as Docker, Singularity and Slurm. All data files, epochs to run, number of replicas and model architectures used for training are controlled by experiment-specific configuration files (e.g. \texttt{string\_cosmic.config}). The network topology is specified in a csv file. Another csv file stores the feature matrix and shows classifications for each node. To run their own experiments, the user only needs to make modifications to the above files. The whole pipeline can then be run by calling Nextflow in the terminal with the desired experiment config file.

\begin{figure}[H]
    \centering
    \includegraphics[width=0.65\textwidth]{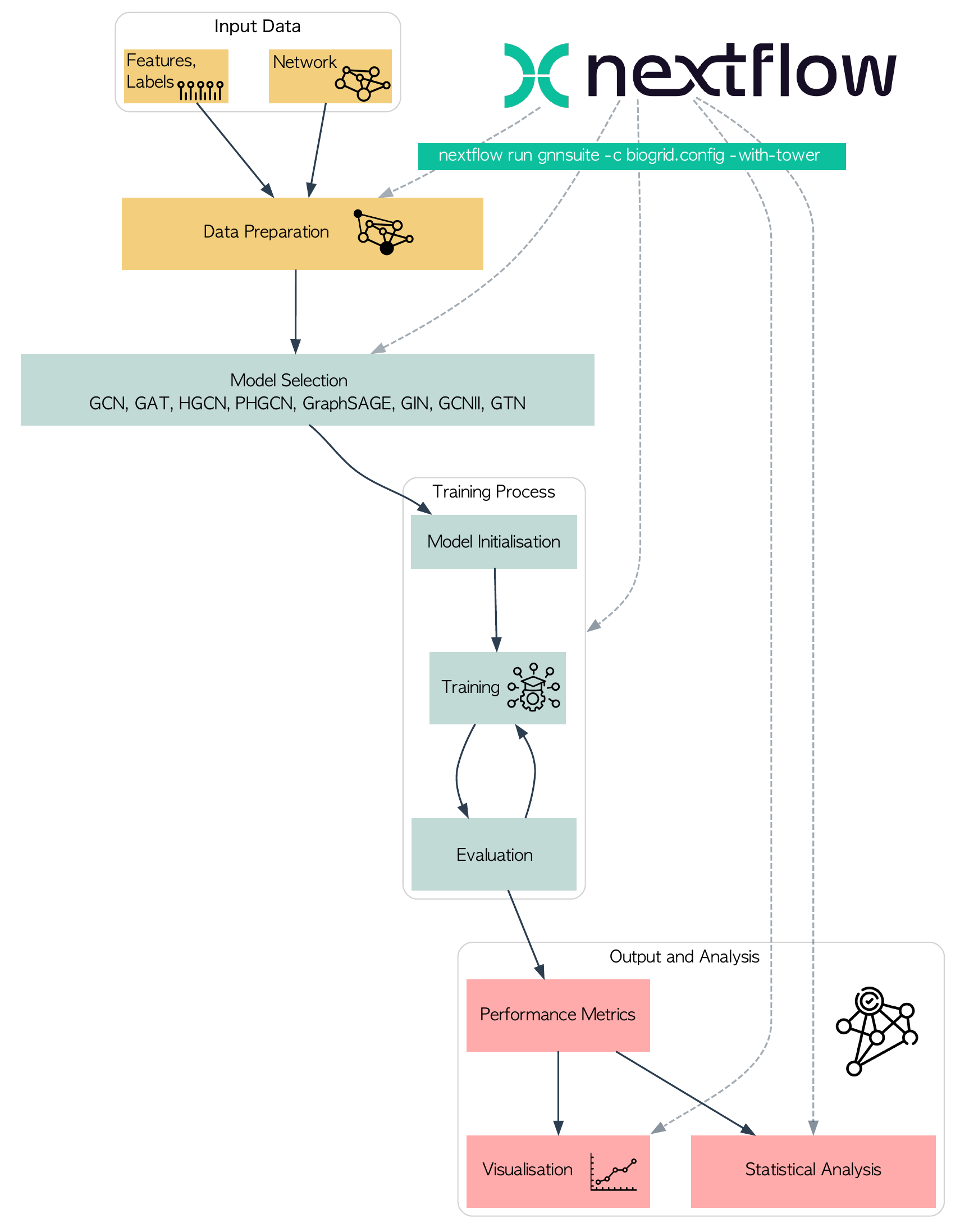}
    \caption{\textbf{GNN-Suite Workflow Diagram}. A flowchart visualising the data pre-processing, training, evaluation, and visualisation steps of the GNN-suite nextflow pipeline as run on the Seqera Platform.}
    \label{fig:nextflow_pipeline}
\end{figure}

The design choice outlined above and shown in Figure \ref{fig:nextflow_pipeline} allows experiments to be easily reproducible and modifiable and to accommodate alternative setups with ease. 

The remaining files of our framework are intended to be static and generally do not need to be modified by the end user. The file \texttt{models.py} defines the GNN architectures we describe in Appendix \ref{app:gnn_architectures} using the PyG library. Further models can be implemented here. The script \texttt{gnn.py} then loads data into a PyG format, builds the model, and computes performance metrics.

We record the following metrics: loss, true negatives (TN), false positives (FP), false negatives (FN), true positives (TP), precision, recall, accuracy, balanced accuracy (BACC) and AUC. In our results, we focus on the BACC value (see below), as the nature of the data in our case study suggested a strong class imbalance. Full details of these metrics are provided in Appendix \ref{appx:evaluationmetrics}.

This pipeline is made accessible via \href{https://github.com/biomedicalinformaticsgroup/gnn-suite}{GNNsuite} a dedicated GitHub repository containing the entire codebase, associated resources, and a detailed user manual. To simplify the setup and create consistent environments, especially for the packages PyTorch (v1.13.1), PyTorch Geometric (v2.3.2) and CUDA (v11.6.1) for GPU computing, we provide a Docker image for the project available via the \href{ghcr.io/stracquadaniolab/gnn-suite:latest}{GitHub Container Registry (GHCR)}. This ensures that the pipeline is easily deployable across various computing platforms.

\subsection*{Data Preparation}
Two different PPI networks were constructed using data derived from either the STRING \cite{szklarczyk_string_2021} or BioGRID \cite{oughtred_biogrid_2021} databases, where nodes represented proteins and edges represented all observed interactions between them. These networks were filtered to retain only interactions labelled as high-confidence within a single connected component, resulting in networks comprising 10,436 nodes and 206,085 edges (STRING) and 5,863 nodes and 16,335 edges (BioGRID). We used the method of Fanfani et al. (2021) \cite{fanfani_discovering_2021} to calculate the likelihood that a node was associated with cancer. To do this, we used $p$-values calculated from the occurrence of recurrent mutations in genomic regions by the Pan-Cancer Analysis of Whole Genomes (PCAWG) project \cite{aaltonen_pan-cancer_2020}. These values estimate the likelihood that an individual region is associated with cancer at the level of coding (CDS), three prime untranslated (3'-UTR), five prime untranslated (5' UTR), promoter, and enhancer regions. We used Fisher's combined probability test under a chi-squared distribution to test the combined significance of the association with cancer at the gene level and used this as a node feature. To complete the data preparation, we required positive labels of known cancer genes. These were obtained from two different cancer panels, the pathway-implicated drivers (PID) \cite{reyna_pathway_2020} and the Catalogue Of Somatic Mutations In Cancer, Cancer Gene Census (COSMIC-CGC) \cite{sondka_cosmic_2018}. Genes identified as cancer drivers in these panels were labelled as $1$ (positive - is a cancer driver gene) and as $0$ (negative - is not a cancer driver gene) otherwise. We used the four different network/label combinations: BioGRID network with COSMIC labels, BioGRID network with PID labels, STRING network with PID labels, STRING network with COSMIC labels to evaluate the ability of different GNN architectures to identify cancer driver genes.

\subsection*{Model Configuration \& Training}
We standardised experiments to establish a benchmark for comparing various convolutional operators and architectures: GCN, HGCN, PHGCN, GAT, GIN, GraphSAGE, GTN, and GCN2 (see Appendix \ref{app:gnn_architectures} for detailed descriptions and equations). This method aims to highlight the strengths and weaknesses of each operator. To ensure a fair comparison, we defined all models with two convolutional layers, a typical design for GNNs, which are generally shallow networks. Note that this constraint means that, on occasion, our models are not exact replicates of the original architectures, since they were adapted in a way that makes them comparable in a common experimental setting.

Each architecture was instantiated with set parameters, including dropout rate and other architecture-specific parameters where applicable. The models were trained using the Adam optimiser, with a learning rate set at 0.01 and a weight decay of $1 \times 10^{-4}$. Given the large class imbalance observed in our datasets, the binary cross-entropy loss function was adjusted with equal weight for positive and negative samples. We used an 80/20 train-test split and trained all models for 300 epochs, which was sufficient to ensure convergence in all cases. For robustness, each model's training was performed 10 times using different random seeds. Dropout was applied after each layer with a value of 0.2 for all architectures. For the GCN2 model, alpha was set to 0.1, and theta was assigned the default none value.

Additionally, we also included a simpler ML model as a baseline. For this purpose, we chose Logistic Regression (LR). Unlike GNN models, where the node representation $h_i^{(l+1)}$ is computed through iterative updates integrating information from neighbouring nodes, LR directly uses the node features $x_i$ to make predictions. 

\clearpage
\section*{Results}
\textbf{Cancer driver gene networks are heterogeneous}

The use of two different sources for both protein-protein interaction data and cancer driver genes allowed us to generate a suite of networks with which to explore sensitivities of different GNN architectures in driver gene classification. As can be seen in Table \ref{tab:network_statistics}, these graphs differ by the size and complexity of the underlying protein–protein interaction networks (PINs) and the node overlap size with the specific gene panel used. This resulted in two classes of networks that differed substantially in size and density. The STRING-derived networks had 1.8x more nodes, 12.6x more edges, a mean node degree 7x higher, and a density 40x higher than the corresponding BioGRID derived networks.

The positive-to-negative ratio (PNR) of cancer driver gene nodes also differed depending on cancer panel source, with nodes having a 1.5x and 1.7x higher PNR in the BioGRID network than the STRING counterpart when labelled with COSMIC and PID data, respectively. Within each network type, those with nodes labelled with COSMIC data had higher PNRs than their corresponding PID-labelled counterparts (BioGRID 3.2x, STRING 3.6x). Together, these statistics emphasised that genomic coverage was broader in STRING versus BioGRID-derived networks, and in COSMIC, more genes are labelled as cancer-driving than in PID. This suite of network configurations allowed us to evaluate how GNN model performance varies across diverse network model structures.

\begin{table}[H]
    \centering
    \resizebox{\textwidth}{!}{
    \begin{tabular}{lccccccc}
        \toprule
        \textbf{Dataset} & \textbf{\#Nodes} & \textbf{\#Edges} & \textbf{Graph Density} & \textbf{Avg Degree} & \textbf{Pos Nodes} & \textbf{Neg Nodes} & \textbf{Pos/Neg Ratio} \\
        \midrule
        STRING PID & 10436 & 206085 & 0.00378 & 39.49 & 177 & 10259 & 0.01725 \\
        STRING COSMIC & 10436 & 206085 & 0.00378 & 39.49 & 615 & 9821 & 0.06262\\
        BioGRID PID & 5863 & 16335 & 0.00095 & 5.57 & 165 & 5698 & 0.02896 \\
        BioGRID COSMIC & 5863 & 16335 & 0.00095 & 5.57 & 498 & 5365 & 0.09282 \\
        \bottomrule
    \end{tabular}
    }
    \caption{\textbf{Cancer driver network statistics}. Key structural and labelling metrics for the four network configurations: STRING–PID, STRING–COSMIC, BioGRID–PID, and BioGRID–COSMIC. Reported metrics include the number of nodes and edges, graph density, average node degree, and the counts of positive and negative cancer driver gene labels.}
    \label{tab:network_statistics}
\end{table}

\textbf{Network-based inference improves cancer driver gene identification}

\begin{figure}
    \begin{figure}[H]
      \includegraphics[width=1\textwidth]{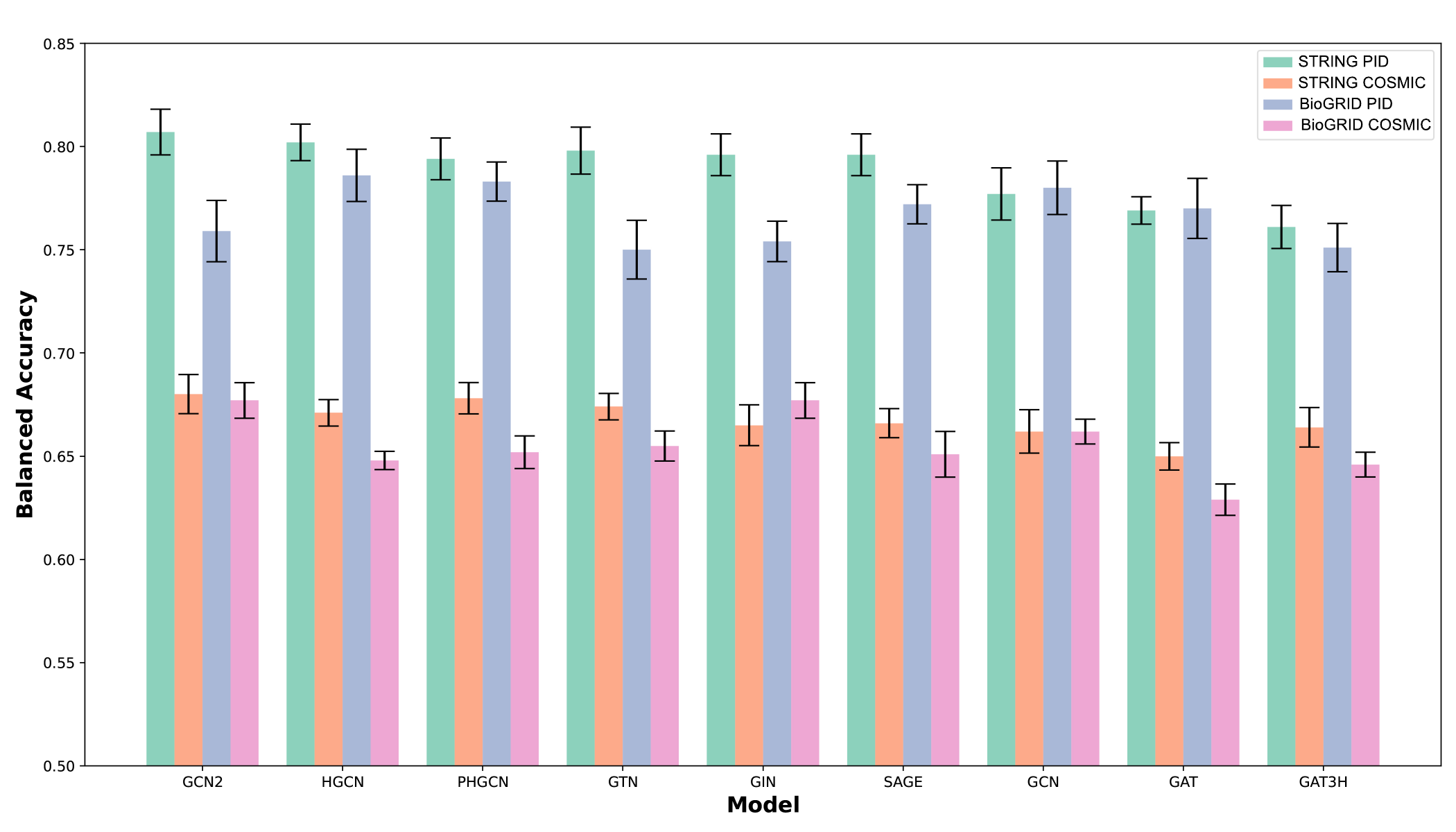}
    \end{figure}
    
    \adjustbox{max width=\textwidth}{
    \begin{tabular}{l|rl|rl|rl|rl}
    \hline &  \multicolumn{2}{c}{STRING PID} &  \multicolumn{2}{c}{STRING COSMIC} &  \multicolumn{2}{c}{BioGRID PID} &  \multicolumn{2}{c}{BioGRID COSMIC}  \\
    \hline
    model &  epoch &    mean ± std &  epoch &    mean ± std &  epoch &    mean ± std &  epoch &    mean ± std  \\
    \hline
    GAT &    185 & 0.769 ± 0.021 & 296 &  0.65 ± 0.021 & 254 &  0.77 ± 0.046  &     84 & 0.629 ± 0.024 \\
    GAT3H &    295 & 0.761 ± 0.033 &    219 &  0.664 ± 0.03 &    287 & 0.751 ± 0.037 & 83 & 0.646 ± 0.019 \\
    GCN &    143 &  0.777 ± 0.04 &     52 & 0.662 ± 0.033 &    285 &  0.78 ± 0.041 &    200 & 0.662 ± 0.019 \\
    GCN2 &    296 & \textbf{0.807 ± 0.035} &    294 &   \textbf{0.68 ± 0.03} &    123 & 0.759 ± 0.047 &    193 & 0.666 ± 0.026 \\
    GIN &    293 & 0.796 ± 0.032 &    295 & 0.665 ± 0.031 &    114 & 0.754 ± 0.031  &     96 & \textbf{0.677 ± 0.027} \\
    GTN &     63 & 0.798 ± 0.036 &     75 &  0.674 ± 0.02 &     60 &  0.75 ± 0.045  &     67 & 0.655 ± 0.023 \\
    HGCN &    243 & 0.802 ± 0.028 &    156 &  0.671 ± 0.02 &    143 &  \textbf{0.786 ± 0.04} &    284 & 0.648 ± 0.014 \\
    PHGCN &    217 & 0.794 ± 0.032 &    184 & 0.678 ± 0.024 &    299 &  0.783 ± 0.03 &    270 & 0.652 ± 0.025 \\
    GraphSAGE &    295 & 0.796 ± 0.032 &    263 & 0.666 ± 0.022 &    256 &  0.772 ± 0.03 &    255 & 0.651 ± 0.035 \\
    Logistic Regression &  ---  & 0.718 ± 0.033 &    ---  & 0.635 ± 0.028 &    ---  & 0.697 ± 0.034 &    ---  & 0.598 ± 0.025 \\
    \hline
    \end{tabular}}
    \caption{\textbf{GNN model benchmarking results}. Models were evaluated on held-out test nodes. The epoch corresponding to the highest mean balanced accuracy (BACC) is reported, with standard deviation calculated across $10$ independent model runs.}
    \label{fig:combined_performance}
\end{figure}

To ensure that we minimise the possibility of bias in model training, we used BACC averaged across 10 independent iterations to evaluate the performance of different GNN model architectures using the four datasets. Figure \ref{fig:combined_performance} shows model performance as the highest mean BACC values alongside the minimum epoch number needed for convergence. For the STRING-PID dataset, the GCN2 model achieved the highest BACC value of 0.807 $\pm$ 0.035, closely followed by HGCN (0.802 $\pm$ 0.028) and GTN (0.798 $\pm$ 0.036). Notably, all models under consideration demonstrated BACC values above 0.75. On the STRING-COSMIC dataset, GCN2 once more outperformed the other models with a BACC of 0.68 $\pm$ 0.03. PHGCN and GTN also performed well, with BACC values of 0.678 $\pm$ 0.024 and 0.674 $\pm$ 0.020, respectively. On the BioGRID dataset, HGCN performed best with a BACC of 0.786 $\pm$ 0.040. This was followed by PHGCN (0.783 $\pm$ 0.030) and GCN (0.78 $\pm$ 0.041). For the BioGRID-COSMIC dataset, GIN achieved the highest value of a BACC (0.677 $\pm$ 0.027). GCN2 and PHGCN also displayed strong performances with values of 0.666 $\pm$ 0.026 and 0.652 $\pm$ 0.025, respectively. Critically, we demonstrate that all network configurations and GNN models demonstrate improved performance over the baseline logistic regression models. The largest performance improvements over baseline for each network are 8.9\% (STRING-PID, GCN2), 4.5\% (STRING-COSMIC, GCN2), 8.9\% (BioGRID-PID, HGCN), and 7.9\% (BioGRID-COSMIC, GIN).

\clearpage

\textbf{GNN model architectures display differential learning efficiency}

\begin{figure}[H]
    \centering
    \includegraphics[width=1\textwidth]{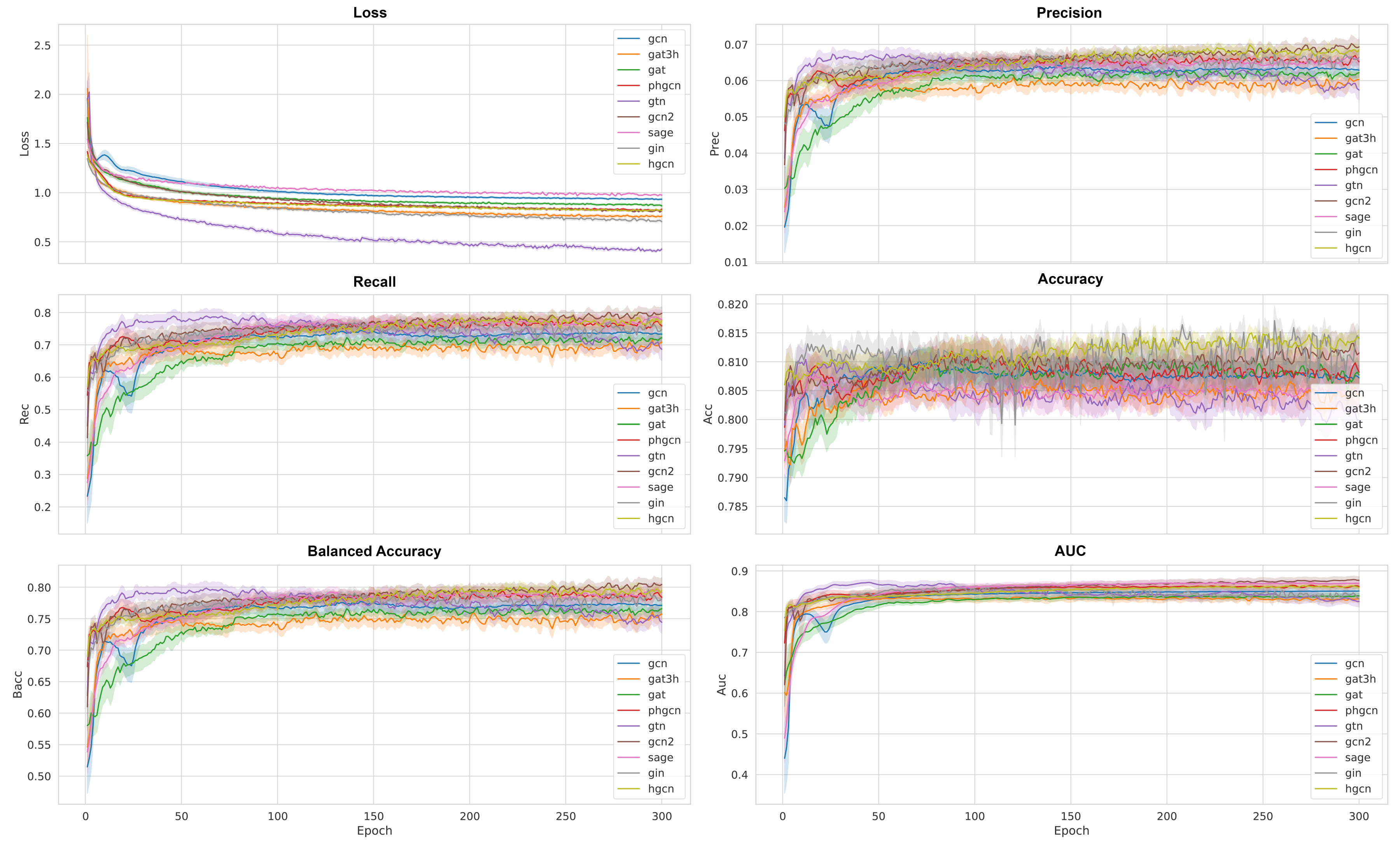}
    \caption{\textbf{Training evaluation metrics}. Variation in training performance metrics for the STRING-PID network configuration for different GNN architectures. Shaded regions around each line represent the standard error over $10$ independent runs.}
    \label{fig:string_model_comp}
\end{figure}

To evaluate GNN model training performance for each network configuration, we captured BACC, loss, and recall across a range of epoch values. In Figure \ref{fig:string_model_comp}, we show these metrics for the STRING PID network with equivalent plots for all other configurations available in Supplementary Figures \ref{fig:string_cosmic_model_comp} (STRING-COSMIC) \ref{fig:biogrid_model_comp} (BioGRID-PID), and \ref{fig:biogrid_cosmic_model_comp} (BioGRID-COSMIC). Most models achieved near-optimal convergence within the first 50 epochs across all network configurations, and final BACC and AUC values that were broadly similar. The GTN model initially converged more rapidly across all metrics (purple lines), suggesting more efficient learning, but BACC began to decline after approximately 80 epochs, indicating possible over-fitting. This behaviour was repeated for GTN models in the other three network configurations and is a known potential issue associated with GTNs due to their high model capacity and susceptibility to lock in to early meta-paths, leading to suboptimal convergence. These data demonstrated the capability of GNN-Suite to efficiently and flexibly compare any number of different GNN configurations across a shared task with common evaluation approaches.

\section*{Discussion}

We introduced GNN-Suite, the first GNN benchmarking framework built using the widely adopted Nextflow scientific workflow system, facilitating reproducibility, portability, and modular design. By using Nextflow, our framework provides an automated, scalable, and user-friendly pipeline for evaluating GNN architectures. It supports a wide range of popular GNN models implemented in PyTorch Geometric (PyG), including GCN, GAT, GraphSAGE, GTN, GIN, GCN2, HGCN, and PHGCN. With its plug-and-play modularity, the suite enables researchers to easily swap in different architectures, datasets, and workflows, making it especially well-suited for those working with complex biological data.

We presented a series of experiments using GNN-Suite to perform a cancer gene driver classification task. Our experiments demonstrated that while a simple logistic regression (LR) model achieved reasonable performance using node features alone, all GNN architectures consistently outperformed it, confirming that incorporating network structure offers predictive advantages. However, when computational resources are limited, LR would be a resource-efficient option for modelling network data. Among the factors that influenced performance, the choice of gene list was most significant. Models trained on the PID gene list consistently outperformed those trained on the COSMIC list. This was surprising as COSMIC is a widely used, manually curated cancer driver panel that is strongly guided by mutational data, continually updated, and often regarded as the gold standard. PID, on the other hand, is largely computationally derived, but crucially incorporates pathway information in addition to mutational signatures. We speculate that it is this alignment between PID data, which uses biological pathway data and PPI data, that makes it particularly well suited to GNN modelling. Several studies have demonstrated that proteins interacting with one another are more likely to operate in common biological pathways and processes \cite{yu_inferring_2012, yuan_analysis_2019}. Interestingly, the underlying PPI network source, whether STRING or BioGRID, had a comparatively minor effect on performance, suggesting that inherent noise in the larger STRING-based dataset was effectively filtered out during training.

While GNN-Suite supports a range of end-to-end functionalities, including data preprocessing, training, evaluation, and basic hyperparameter tuning via Optuna, it currently focuses on binary node classification tasks. Future work will extend the framework to support multi-class classification, link prediction, and graph classification. Additionally, while the current version uses PyG exclusively, we aim to incorporate compatibility with other frameworks such as TensorFlow-GNN and DGL in future updates.

We believe the modularity, reproducibility, and ease of use of GNN-Suite are key enablers for researchers. We demonstrated how it can be easily used to perform thorough benchmarking of GNN model performance across a diverse set of biological datasets, including those from PCAWG, COSMIC, BioGRID, PID, and STRING and to draw meaningful conclusions in cancer gene classification tasks.

We hope that GNN-Suite will help democratise access to advanced GNN techniques for biomedical researchers, enabling them to conduct more rigorous and efficient model evaluations on domain-specific datasets. Ultimately, we envision GNN-Suite supporting a broader community in unlocking new insights from complex biological networks.

\bibliographystyle{plain} 
\bibliography{main} 

\section*{Acknowledgements}
\label{sec:acknowledgements}

This work was supported by the United Kingdom Research and Innovation (grant EP/S02431X/1), UKRI Centre for Doctoral Training in Biomedical AI at the University of Edinburgh, School of Informatics. For the purpose of open access, the author has applied a Creative Commons Attribution (CC BY) licence to any author accepted manuscript version arising. 

\subsection*{Author contributions statement}
SK designed the study, developed the code, conducted experiments, and drafted the manuscript. GS and IS contributed to designing the project, providing overall supervision and discussions. GS and IS revised the manuscript and approved the final version.
Special thanks to Dominic Phillips for proofreading this article.
\newpage

\appendix

\renewcommand{\thesubsection}{S\arabic{subsection}}

\renewcommand{\thefigure}{S\arabic{figure}}
\setcounter{figure}{0}

\section*{Supplementary Material}
\subsection{Graph Neural Networks (GNNs)}
\label{app:gnns}

Graph Neural Networks (GNNs) are a specialised class of neural networks tailored for graph-structured data, commonly used for tasks such as node classification, graph classification, and link prediction \cite{velickovic_message_2022}. They stand out as one of the most adaptable deep learning (DL) architectures, with many other DL models being interpretable as specific cases of GNNs. 

A graph $G = (V, E)$, is a tuple of a node set $V$ and an edge set $E$. Nodes represent entities of interest, while edges capture the relationships or interactions between these entities. Graphs can be directed, undirected, or weighted. The adjacency matrix, symbolised as $A$, represents the connections. For an undirected, unweighted graph with $n$ nodes, $A$ is an $n \times n$ matrix where $A_{ij}$ is 1 if nodes $i$ and $j$ are connected, and 0 otherwise. For a directed graph, $A_{ij} = 1$ if and only if there is a directed edge from $i$ to $j$. For weighted graphs, the value of $A_{ij} \in \mathbb{R}^+$  indicates the strength of the connections. In the context of GNN training, each node $v_i \in V$ of the graph is associated with a feature vector $x_i$. Similarly, each edge $(v_i, v_j) \in E$ may be linked with a weight or label $w_{ij}$ \cite{bronstein_geometric_2021}.\\

\textbf{Message Passing}

GNNs use the concept of \textit{message passing}. This is an iterative process, where the hidden state of a node aggregates information from an increasingly extended neighbourhood of surrounding nodes. At the start of message passing, the hidden state is initialised with the feature vector: \(h_i^{(0)} = x_i\) \cite{gilmer_neural_2017}. In each subsequent iteration, the hidden state is updated by aggregating the hidden states of neighbouring nodes. As message passing proceeds, the GNN aggregates information from a node's extended neighbourhood, capturing local graph structure. This process refines the node features into a representation that encapsulates the node's role within its neighbourhood. The general update rule is of the form:
\begin{equation}
     h_i^{(l+1)} = \phi \left( h_i^{(l)}, \bigoplus_{j \in \mathcal{N}(i)} \psi(h_i^{(l)}, h_j^{(l)}) \right) 
     \label{eq:gnn_node_rep}
\end{equation}

where \( h_i^{(l)} \) is the hidden state of node \( i \) after $l$ iterations, \( \mathcal{N}(i) \) denotes the set of neighbours of node \( i \), $\psi$ is a parameterised function that combines two hidden states (e.g. linear transformation, edge-conditioned function, attention mechanism or neural network), \( \bigoplus \) is an aggregation function (typically sum, mean or max operation), and $\phi$ is another parameterised function. 

As with other DL frameworks, the parameters of the functions $\psi$ and $\phi$ are updated during training via gradient descent on a suitable loss function, such as mean squared error (for regression tasks) and cross-entropy and binary cross-entropy loss (for classification tasks). 

In the rapidly evolving field of ML, there has been a push towards developing more standardised frameworks for rigorously comparing model architectures on common tasks \cite{olson_pmlb_2017, hu_open_2021, thiyagalingam_scientific_2022, greener_guide_2022}. Such comparisons are valuable because they make benchmarking more streamlined, help highlight potential drawbacks and failure modes of existing architectures, and provide a blueprint framework for subsequent research, enabling wider community adoption of state-of-the-art methods \cite{munafo_manifesto_2017, wilkinson_fair_2016, vicente-saez_open_2018, pineau_improving_2021}.

Within the GNN domain in particular, many of the most impactful scientific applications are in biology and biology-adjacent fields \cite{stokes_deep_2020, ahmedt-aristizabal_survey_2022, schulte-sasse_integration_2021, zhang_graph_2021, rong_self-supervised_2020}. Although several benchmarking frameworks have been developed \cite{dwivedi_benchmarking_2022, gong_gnnbench_2024}, none of them explicitly use frameworks that are widely adopted in the biological sciences. There is therefore a clear community need to develop a flexible GNN framework with a minimal learning curve for computational biologists.

\subsection{Nextflow}
\label{app:nextflow}

Nextflow is a workflow management system designed to support reproducible data analysis pipelines \cite{di_tommaso_nextflow_2017}. It enables scalable and portable scientific workflows across a variety of computational platforms, including local machines, HPC clusters, and cloud environments. It is implemented in Groovy and Java \cite{jackson_using_2020} and employs a programming model in which processes communicate through channels, facilitating parallel and distributed execution. It allows for automatic failure recovery, and experiments can be monitored through the Seqera Platform. Nextflow is widely used in bioinformatics and other data-intensive disciplines for automating complex data processing tasks and enhancing reproducibility \cite{langer_empowering_2024}. As of 2023, Nextflow is downloaded by over $120,000$ users per month and continues to have an expanding user base \cite{ortiz2023reflecting}. 

\subsection{GNN Architectures}
\label{app:gnn_architectures}

To better understand the behaviour of different GNN architectures, we focused on a selection of eight GNN architectures implemented in PyTorch Geometric (PyG) \cite{fey_fast_2019}. This section provides an overview of the GNN architectures included in our study: Graph Convolutional Networks (GCN), Graph Attention Networks (GAT), Hierarchical Graph Convolutional Networks (HGCN), Parallel Hierarchical Graph Convolutional Networks (PHGCN), GraphSAGE, Graph Isomorphism Network (GIN), Graph Convolutional Network II (GCN2), and Graph Transformer Network (GTN).

\subsubsection{Graph Convolutional Network (GCN)}

The GCN is the simplest operator in GNNs that performs a simple convolution over the graph. It leverages spectral graph theory to perform convolution operations on graphs. GCN aims to learn node representations by aggregating information from their local neighbourhoods.
It has been widely used for various tasks such as node classification, link prediction, and graph classification \cite{kipf_semi-supervised_2017}.

In the general GNN framework, \( h_i^{(l+1)} \) represents the updated node representation at layer \( l+1 \). This representation is often derived from initial node features, denoted as \( x_i \). As information progresses through the layers of the network, the initial features \( x_i \) are transformed and enriched by aggregating information from neighbouring nodes, leading to the refined representations \( h_i^{(l+1)} \). In the context of GCNs, this transformation is achieved through a specialised convolutional operator, designed to capture both local structure and global patterns in the graph.

The convolutional operator used in GCN is given by:
\begin{equation}
\label{eq:gcn-conv}
    H^{(l+1)} = \sigma \left(\tilde{D}^{-\frac{1}{2}} \tilde{A} \tilde{D}^{-\frac{1}{2}} H^{(l)} W^{(l)} \right),
\end{equation}

where ${H}^{(l)}$ is the node representation matrix of the \(l\)th layer, whose \(i\)th row is equal to the node representation of the \(i\)th node in that layer, \( h_i^{(l)} \). \(\tilde{A} = A + I\) denotes the adjacency matrix with inserted self-loops, \(\tilde{D}_{ii} = \sum_{j=0} \tilde{A}_{ij}\) represents its diagonal degree matrix of $\tilde{A}$, $W^{(l)}$ is the weight matrix in the \(l\)th layer, and $\sigma$ is a non-linear activation function.

The node-wise formulation of the GCN operator is:
\begin{equation}
    h^{(l+1)}_i = \sigma \left(W^{(l)\top} \sum_{j \in \mathcal{N}(i) \cup \{ i \}} \frac{e_{j,i}}{\sqrt{\tilde{d}_j \tilde{d}_i}} h_j^{(l)} \right)
\end{equation}
where \(\tilde{d}_i = 1 + \sum_{j \in \mathcal{N}(i)} e_{j,i}\), where \(e_{j,i}\) denotes the edge weight from source node \(j\) to target node \(i\).

\subsubsection{Hierarchical Graph Convolutional Networks (HGCN)}
We define a Hierarchical Graph Convolutional Network (HGCN) as a multi-layer GCN with a final layer that aggregates information from all intermediate layers via concatenation.
Instead of passing the output of one layer directly as the input to the next, HGCN maintains a list of outputs from all layers (layer-wise output concatenation). These outputs are then concatenated along the feature dimension before being passed through a final linear layer to predict the class labels. This can be represented as:
\begin{equation}
    z = W_{\text{final}} \cdot \text{concat}(H^{(1)}, H^{(2)}, \ldots, H^{(L)}),
\end{equation}

where $L$ is the number of layers, $W_{\text{final}}$ is a weight matrix with learnable parameters and $z$ is the final output.
$H^{(1)}, H^{(2)} \ldots , H^{(L)}$ are the outputs of GCN convolutional layers (see \eqref{eq:gcn-conv}).

\subsubsection{Parallel Hierarchical Graph Convolutional Networks (PHGCN)}
PHGCN is a modification of HGCN that places the GCN convolutional layers parallel rather than in series; that is,

\begin{equation}
    z = W_{\text{final}} \cdot \text{concat}(H_{1}, H_{2}, \ldots, H_{L}),
\end{equation}
where
\begin{equation}
    H_{l}=\sigma \left(\tilde{D}^{-\frac{1}{2}} \tilde{A} \tilde{D}^{-\frac{1}{2}} H^{(0)} W^{(l)} \right)
\end{equation}

and where $L$ is the number of layers, $W_{\text{final}}$ is a weight matrix with learnable parameters and $z$ is the final output.
$H_{1}, H_{2} \ldots, H_{L}$ are the outputs of GCN convolutional layers with the same original input $H^{(0)}$ (see \eqref{eq:gcn-conv}).

\subsubsection{Graph Attention Networks (GAT)}

The Graph Attention Network (GAT) uses an attention mechanism that allows nodes to assign varying importance weights to their neighbours during message aggregation. In this study, we additionally include a variant of this model with three attention heads and refer to it as GAT3H \cite{velickovic_graph_2018}.

The convolutional operator in GAT is defined as:
\begin{equation}
    h^{(l+1)}_i = \sigma\left(\alpha_{i,i}W^{(l)}h^{(l)}_{i} + \sum_{j \in \mathcal{N}(i)} \alpha_{i,j}W^{(l)}h^{(l)}_{j}\right)
\end{equation}

Where the attention coefficients \(\alpha_{i,j}\) are computed as:
\begin{equation}
    \alpha_{i,j} = \frac{\exp\left(\mathrm{LeakyReLU}\left(a^{\top} [W^{(l)}h^{(l)}_i \, \Vert \, W^{(l)}h^{(l)}_j]\right)\right)}{\sum_{k \in \mathcal{N}(i) \cup \{ i \}} \exp\left(\mathrm{LeakyReLU}\left(a^{\top} [W^{(l)}h^{(l)}_i \, \Vert \, W^{(l)}h^{(l)}_k]\right)\right)},
\end{equation}

where $a$ is a learnable weight vector for the attention mechanism, and $||$ denotes vector concatenation.

\subsubsection{GraphSAGE}

GraphSAGE extends the idea of the GCN by introducing a sampling-based aggregation mechanism. Unlike GCN and GAT, which operate on the entire neighbourhood or apply attention mechanisms, GraphSAGE samples a fixed-size neighbourhood and aggregates the features of sampled nodes. This inductive learning approach allows it to generalise to unseen nodes, making it well-suited for large-scale graphs \cite{hamilton_inductive_2018}. 

The convolutional operator in GraphSAGE is:
\begin{equation}
h^{(l+1)}_i = \sigma\left(W_1^{(l)}\cdot h_i^{(l)} + W_2^{(l)}\cdot \mathrm{mean} \left(\{h^{(l)}_j,\; \forall j \in \mathcal{N}(i) \} \right)\right),
\end{equation}

where the mean function is applied element-wise across the feature vectors. This approach allows GraphSAGE to capture local neighbourhood information effectively while remaining scalable to large networks \cite{hamilton_inductive_2018}.

\subsubsection{Graph Isomorphism Network (GIN)}

GIN is designed to capture the graph structure in a way that distinguishes different graph topologies, making it powerful for tasks requiring sensitivity to structural variation. Unlike GCN, GAT, and GraphSAGE, which focus on local neighbourhood aggregation through different mechanisms, GIN aggregates features from neighbours while considering the global graph structure, ensuring that different graph structures lead to different embeddings \cite{xu_how_2019}.

The convolutional operator in GIN is defined as:
\begin{equation}
h^{(l+1)}_i = h_{\theta} \left( (1 + \epsilon) \cdot h^{(l)}_i + \sum_{j \in \mathcal{N}(i)} h^{(l)}_j \right)
\end{equation}
where $h_{\theta}$ denotes a neural network (NN), typically a multi-layer perceptron (MLP), and $\epsilon$ is a scalar that can be either fixed or learned. In our experiments, we set $\epsilon = 0$.

\subsubsection{GCN2}

GCN2 builds upon the original GCN by introducing initial residual connections and identity mapping. It addresses the often encountered issue in deeper GNN architectures, called over-smoothing. It achieves this by incorporating the initial feature representation and identity mapping via residual connections. This makes GCN2 suitable for deeper architectures \cite{chen_simple_2020}.

The convolutional operator in GCN2 is given by:
\begin{equation}
    H^{(l+1)} = \sigma \left( \left( (1 - \alpha^{(l)}) \tilde{D}^{-\frac{1}{2}} \tilde{A} \tilde{D}^{-\frac{1}{2}}H^{(l)} + \alpha^{(l)} H^{(0)}\right) \left( (1 - \beta^{(l)}) I + \beta^{(l)} W^{(l)} \right) \right),
\end{equation}

where $H^{(0)}$ is the initial feature representation, $\alpha$ models the strength of the initial residual connection to the input features, and $\beta$ denotes the strength of the identity mapping.

\subsubsection{Transformer Conv}

The \texttt{TransformerConv} class introduces a graph transformer operator inspired by the transformer architecture, specifically adapted for graph-structured data. This operator is derived from the paper ``Masked Label Prediction: Unified Message Passing Model for Semi-Supervised Classification'' \cite{shi_masked_2021}.

The convolutional operator is defined as:
\begin{equation}
    h_i^{(l+1)} = \sigma \left( W_1^{(l)} h_i^{(l)} + \sum_{j \in \mathcal{N}(i)} \alpha^{(l)}_{i,j} W_2^{(l)} h_{j}^{(l)} \right),
\end{equation}
where the attention coefficients, $\alpha_{i,j}$, are computed using multi-head dot product attention:
\begin{equation}
\alpha^{(l)}_{i,j} = \textrm{softmax} \left( \frac{(W^{(l)}_3h_i^{(l)})^{\top} (W^{(l)}_4h_j^{(l)})}{\sqrt{{d}^{(l)}}} \right)
\end{equation}

and where $d^{(l)}= \text{dim}(h^{(l)})$ denotes the hidden dimensionality of each attention head.

A notable feature of the \texttt{TransformerConv} class is its ability to incorporate edge features into the convolutional operator,  enhancing the model’s ability to capture edge-specific information. When edge features are present, the convolutional operator is modified accordingly.

This class demonstrates the growing potential of transformer architectures in GDL tasks.

\clearpage

\subsection{Training Progress}
\label{app:training_progress}
\subsubsection{STRING-COSMIC}

\begin{figure}[H]
    \centering
    \includegraphics[width=1\textwidth]{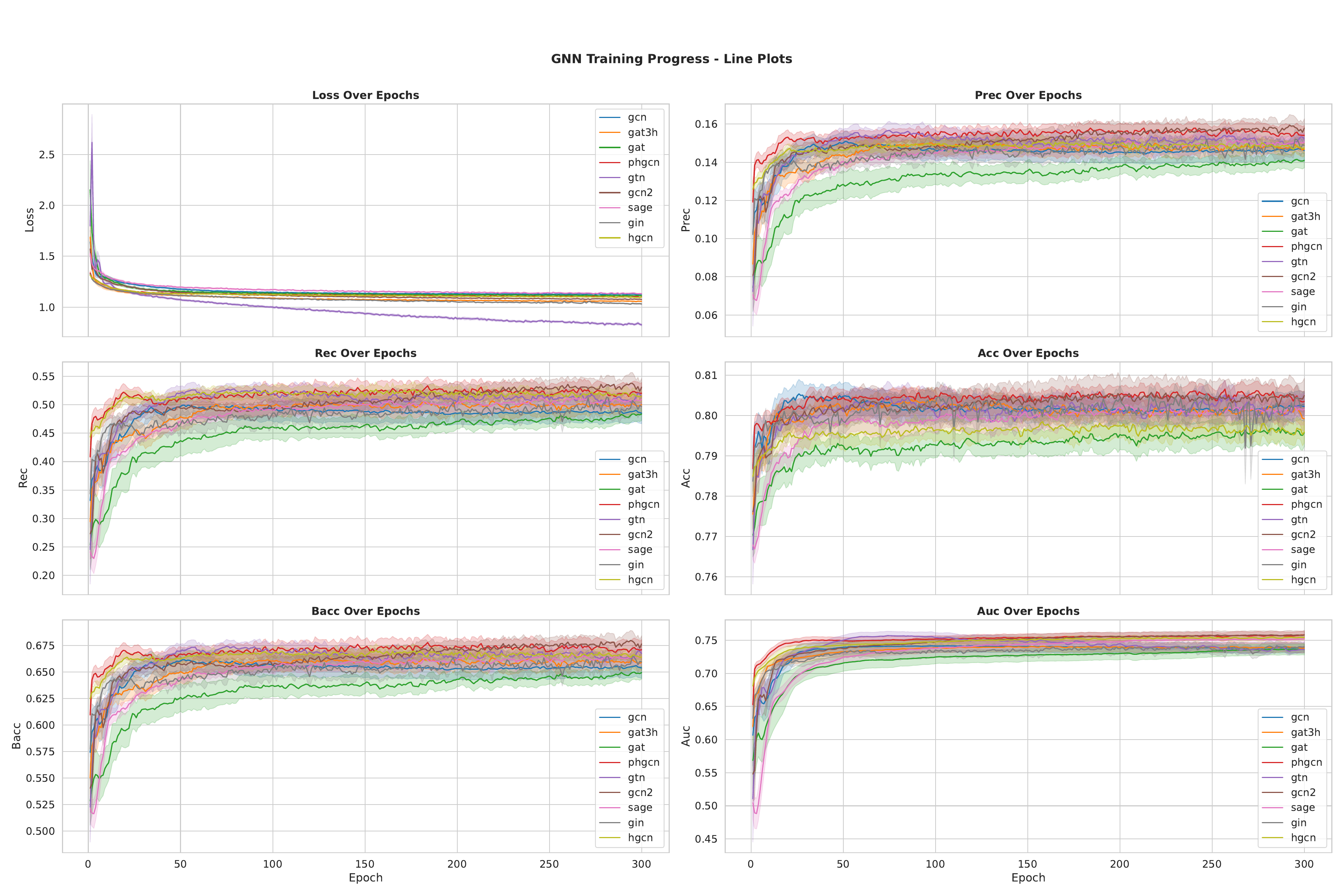}
    \caption{Evaluation metrics on the test set during training, including the (training) loss on the STRING-COSMIC dataset for different GNN architectures. Shaded regions around each line represent the standard error over $10$ independent runs.}
    \label{fig:string_cosmic_model_comp}
\end{figure}

\clearpage

\subsubsection{BioGRID-PID}

\begin{figure}[H]
    \centering
    \includegraphics[width=1\textwidth]{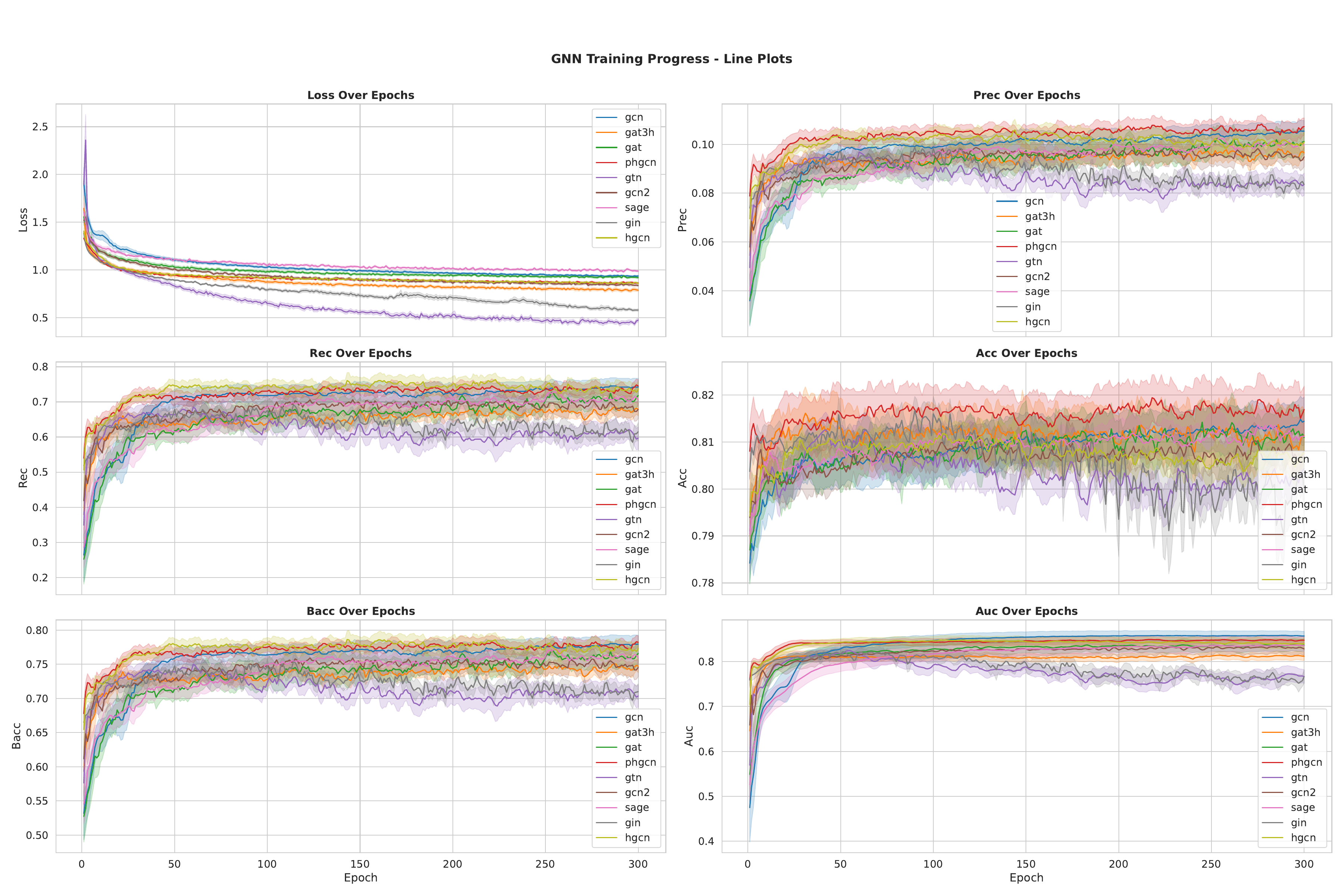}
    \caption{Evaluation metrics on the test set during training, including the (training) loss on the BioGRID-PID dataset for different GNN architectures. Shaded regions around each line represent the standard error over $10$ independent runs.}
    \label{fig:biogrid_model_comp}
\end{figure}

\clearpage

\subsubsection{BioGRID-COSMIC}

\begin{figure}[H]
    \centering
    \includegraphics[width=1\textwidth]{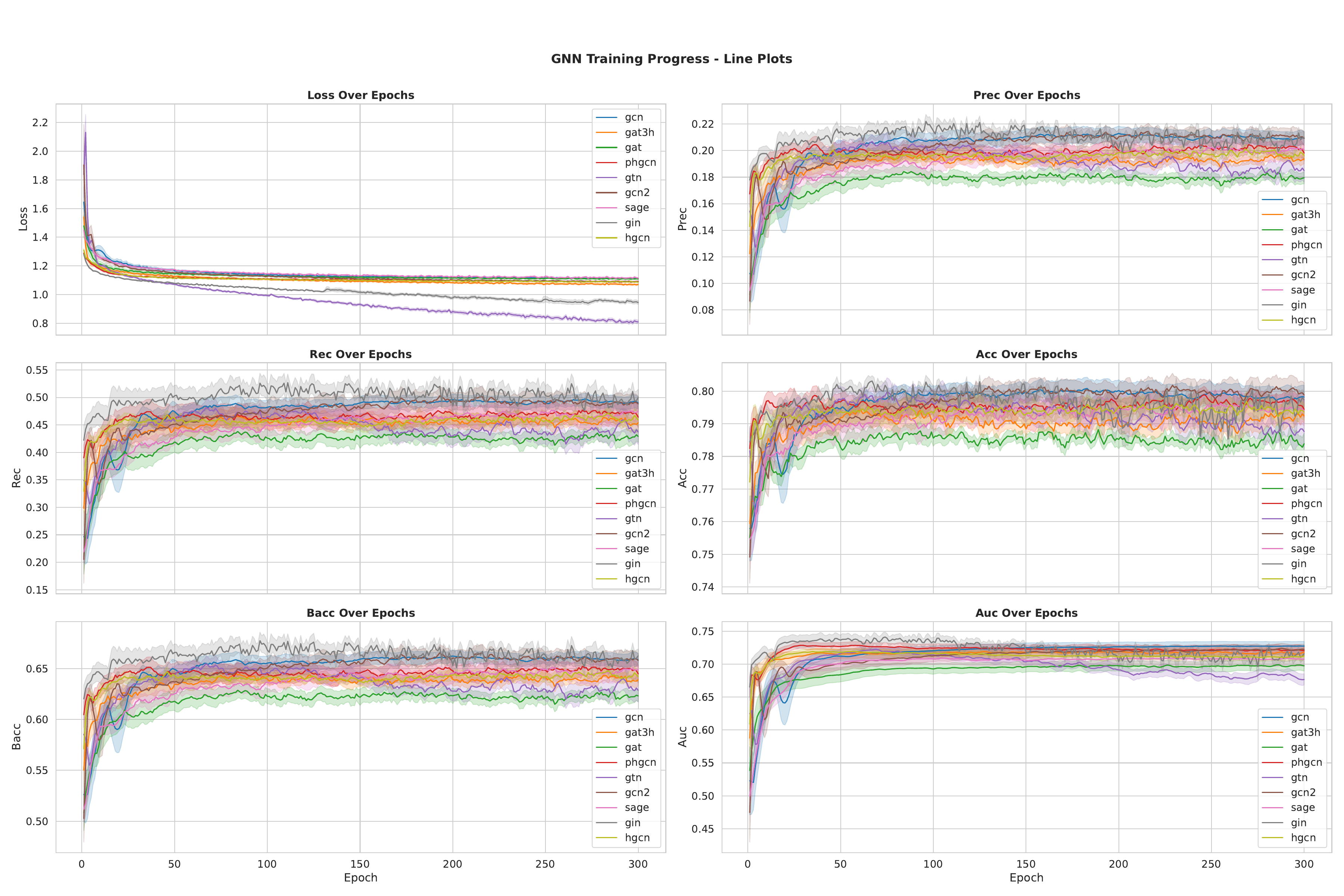}
    \caption{Evaluation metrics on the test set during training, including the (training) loss on the BioGRID-COSMIC dataset for different GNN architectures. Shaded regions around each line represent the standard error over $10$ independent runs.}
    \label{fig:biogrid_cosmic_model_comp}
\end{figure}

\clearpage

\subsection{Evaluation metrics}
\label{appx:evaluationmetrics}

In our evaluation process, we utilise a set of standard evaluation metrics for binary classification. Outcomes are typically labelled as either positive or negative, and model performance is assessed by comparing its predictions to the actual outcomes, using components of the confusion matrix: true positives (TPs), false positives (FPs), true negatives (TNs), and false negatives (FNs). These metrics form the foundation of our evaluation approach, as outlined by Berrar \cite{berrar_performance_2019}.

While classification accuracy, which is the proportion of correctly predicted instances over the total number of instances, is a commonly used metric, it can be less informative in cases with imbalanced class distributions. For instance, if a classifier frequently identifies instances as the majority class, the accuracy might not fully represent its performance, especially if its predictions for the minority class are not accurate.

\begin{equation}
    Accuracy = \frac{TP+TN}{TP+TN+FP+FN}
\end{equation}
To address this, we also consider balanced accuracy (BACC). This metric adjusts for class imbalance by assigning weights to instances based on the inverse frequency of their actual class.
\begin{equation}
     \text{Balanced-Accuracy} = \frac{1}{2}\left(\frac{TP}{TP+FN}+\frac{TN}{TN+FP}\right)
\end{equation}

For binary classification tasks, BACC is equivalent to the average of sensitivity and specificity\footnote{\href{https://scikit-learn.org/stable/modules/model_evaluation.html}{Scikit-learn User Guide}}.

Precision is often referred to as the positive predictive value. It indicates the proportion of correctly predicted positive instances to all positive predictions.
\begin{equation}
    \text{Precision} = \frac{TP}{TP+FP}
\end{equation}
Recall, also known as sensitivity, represents the proportion of actual positives that were correctly identified by the model.
\begin{equation}
    \text{Recall} = \frac{TP}{TP+FN}
\end{equation}

Additionally, the Area Under the Curve (AUC) is employed to quantify the model's discriminative ability between positive and negative classes. Specifically, AUC measures the area under the Receiver Operating Characteristic (ROC) curve, which plots the true positive rate (TPR) against the false positive rate (FPR) for various thresholds. 

\begin{equation}
    \text{AUC} = \int_0^1 \text{True Positive Rate} , d(\text{False Positive Rate})
\end{equation}

This metric, in conjunction with others, offers a comprehensive view of the model's classification capabilities.

We also monitor the change in loss during training. In DL, the loss function serves as a guide, directing the model towards better performance. A declining loss shows that the model is effectively learning, while a plateau or an increase could suggest potential overfitting or that the model has achieved its optimal performance. The trajectory of the loss curve can also show how optimal certain hyperparameters are, such as the learning rate. By monitoring both the training and validation loss, we can select the optimal moment to stop training, and so prevent overfitting (early stopping). Additionally, when assessing various GNN architectures, a side-by-side comparison of their respective loss curves can be informative of their relative stability and rate of convergence.

\newpage

\end{document}